%% file: apsipa2021_ito-hiroki.tex
\documentclass[conference,a4paper]{APSIPA2021}
\usepackage[pdftex]{graphicx}
\usepackage{hyperref}
\usepackage{cite}
\usepackage{multirow}
\usepackage{amsmath}

\let\labelindent\relax
\usepackage{enumitem}

\begin{document}

\title{Access Control Using Spatially Invariant Permutation of Feature Maps for Semantic Segmentation Models}

\author{%
\authorblockN{%
Hiroki Ito,
MaungMaung AprilPyone, and
Hitoshi Kiya
}
\authorblockA{Tokyo Metropolitan University, Japan}
}

\maketitle
\thispagestyle{empty}

\begin{abstract}
\input ./sections/abstract.tex
\end{abstract}

\section{Introduction}
\label{sec:intro}
\input ./sections/intro.tex


\section{Proposed Method}
\label{sec:proposed}
\input ./sections/proposed.tex

\section{Experiments and Results}
\label{sec:experiment}
\input ./sections/experiment.tex

\section{Conclusion}
\label{sec:conclusion}
\input ./sections/conclusion.tex

\section*{Acknowledgment}
This study was partially supported by JSPS KAKENHI (Grant Number JP21H01327) and Support Center for Advanced Telecommunications Technology Research, Foundation (SCAT).

\bibliographystyle{IEEEtran}
\bibliography{strings}

\end{document}

%% file: sections/abstract.tex
In this paper, we propose an access control method that uses the spatially invariant permutation of feature maps with a secret key for protecting semantic segmentation models.
Segmentation models are trained and tested by permuting selected feature maps with a secret key.
The proposed method allows rightful users with the correct key not only to access a model to full capacity but also to degrade the performance for unauthorized users.
Conventional access control methods have focused only on image classification tasks, and these methods have never been applied to semantic segmentation tasks.
In an experiment, the protected models were demonstrated to allow rightful users to obtain almost the same performance as that of non-protected models but also to be robust against access by unauthorized users without a key.
In addition, a conventional method with block-wise transformations was also verified to have degraded performance under semantic segmentation models.

%% file: sections/intro.tex
Deep neural networks (DNNs) have led to major breakthroughs in computer vision for a wide range of applications.
Convolutional neural networks (CNNs) are a type of DNN.
Current commercial applications for image recognition, object detection, and semantic segmentation are primarily powered by CNNs \cite{dl2, application0}.
Therefore, CNNs have become the de facto standard for visual recognition systems for many different applications. 
However, training successful CNNs requires three ingredients: a huge amount of data, GPU-accelerated computing resources, and efficient algorithms, and this is not a trivial task. 
Therefore, trained CNNs have great business value. 
Considering the expenses necessary for the expertise, money, and time taken to train a CNN model, a model should be regarded as a kind of intellectual property (IP).

There are two aspects of IP protection for DNN models: ownership verification and access control. 
The former focuses on identifying the ownership of models, and the latter addresses protecting the functionality of DNN models from unauthorized access. 
Ownership verification methods are inspired by digital watermarking, and they embed watermarks into DNN models so that the embedded watermarks can be used to verify the ownership of the models in question \cite{embed0, embed1, embed2, embed3, embed4, embed5, embed6, embed7, embed8}.

Although the above watermarking methods can facilitate the identification of the ownership of models, in reality, a stolen model can be exploited in many different ways. 
For example, an attacker can use a model for their own benefit without arousing suspicion, or a stolen model can be used for model inversion attacks \cite{inv-attack} and adversarial attacks \cite{adv1}.
Therefore, it is crucial to investigate mechanisms for protecting DNN models from unauthorized access and misuse. 
In this paper, we focus on protecting a model from misuse when it has been stolen (i.e., access control).

A method for protecting a model against unauthorized access was inspired by adversarial examples \cite{adv1, adv2, adv3} and image encryption \cite{enc1, enc2, enc3, enc4, perceptual1, perceptual3}, and it was proposed to utilize secret perturbation to control the access of a model \cite{access1}.
Another study introduced a secret key for protecting a model \cite{access2, access3}. 
The secret key-based protection method \cite{access3} uses a key-based transformation that was originally used by an adversarial defense in \cite{adv-def}, which was in turn inspired by perceptual image encryption methods \cite{perceptual1, perceptual2, perceptual3, perceptual4, perceptual5, perceptual6, perceptual7, perceptual8}. 
This model protection method utilizes a secret key in such a way that a stolen model cannot be used to its full capacity without a correct secret key. 

However, these methods were evaluated only on image classification tasks, and it is not known how well they perform on other advanced tasks. 
Therefore, in this paper, we consider protecting semantic segmentation models from unauthorized access for the first time, and we propose a novel access control method that uses the spatially invariant permutation of feature maps on the basis of a secret key. 
The proposed method allows rightful users with the correct key to access a model to full capacity and degrade the performance for unauthorized users. 
In experiments on semantic segmentation, we evaluated the access control performance of models trained by using feature map permutation. 
The results show that the protected models provided almost the same segmentation performance as that of non-protected models against authorized access, while the segmentation accuracy seriously dropped when an incorrect key was given. 
Furthermore, the conventional method proposed for image classification tasks \cite{access3} was demonstrated to have degraded performance under semantic segmentation tasks.

%% file: sections/proposed.tex
\subsection{Overview}
\begin{figure}[t]
	\centering
 	\centerline{\includegraphics[width=\linewidth] {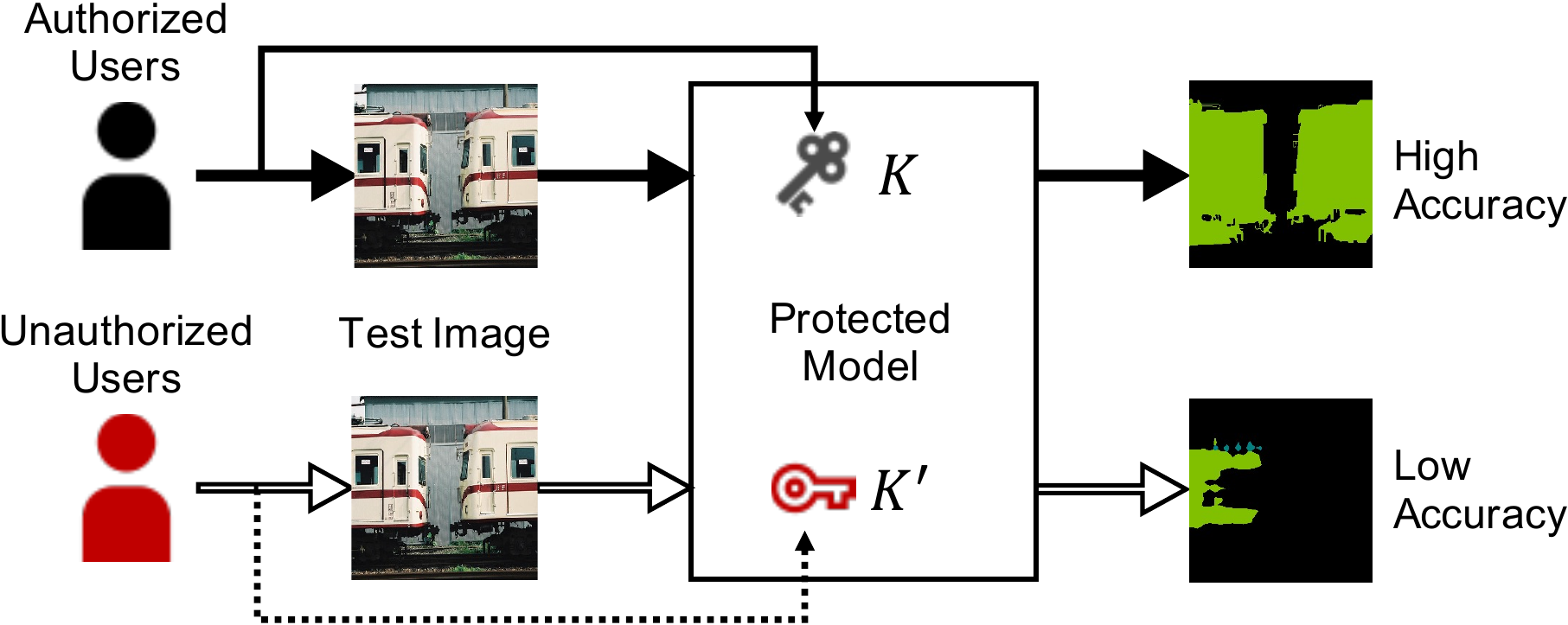}}
  	\caption{Overview of access control}
	\label{fig:framework}
\end{figure}

Figure \ref{fig:framework} illustrates an overview of access control for protecting semantic segmentation models from unauthorized access.
A protected model is prepared by training a network with secret key $K$.
Authorized users input test images into the protected model with correct key $K$, in which the model provides almost the same segmentation map as that predicted by using a non-protected model.
In contrast, when unauthorized users who do not know key $K$ input test images into a protected model without any key or with incorrect (estimated) key $K'$, the model provides a degraded map.

\subsection{Semantic Segmentation}
\begin{figure}[t]
	\centering
 	\centerline{\includegraphics[width=\linewidth] {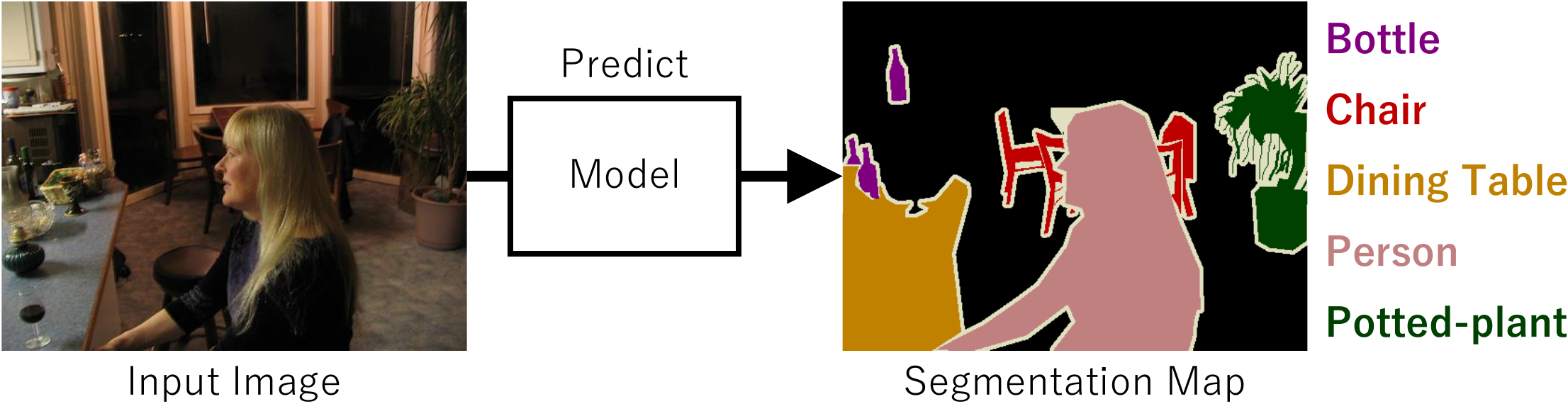}}
  	\caption{Example of semantic segmentation}
	\label{fig:semantic}
\end{figure}

The goal of semantic segmentation is to understand what is in an image at the pixel level.
Figure \ref{fig:semantic} shows an example of semantic segmentation.
The segmentation model predicts a segmentation map from an input image, where each pixel in the segmentation map represents a class label.

The mean intersection-over-union (mean IoU) \cite{giou, fcn} is used as a metric for evaluating the segmentation performance.
The mean IoU value of an image is calculated by averaging the IoU of each class, which is defined by
\begin{equation}
    \label{eq:iou}
    \mathrm{IoU} = \frac{TP}{TP+FP+FN},
\end{equation}
where $TP$, $FP$, and $FN$ mean true positive, false positive, and false negative values calculated from a predicted segmentation map and ground truth one, respectively.
In addition, the metric ranges from zero to one, where a value of one means that a predicted segmentation map is the same as that of the ground truth, and a value of zero indicates they have no overlap.

\subsection{Training Model with Key \texorpdfstring{$K$}{K}}
\begin{figure*}[t]
	\centering
 	\centerline{\includegraphics[width=\linewidth] {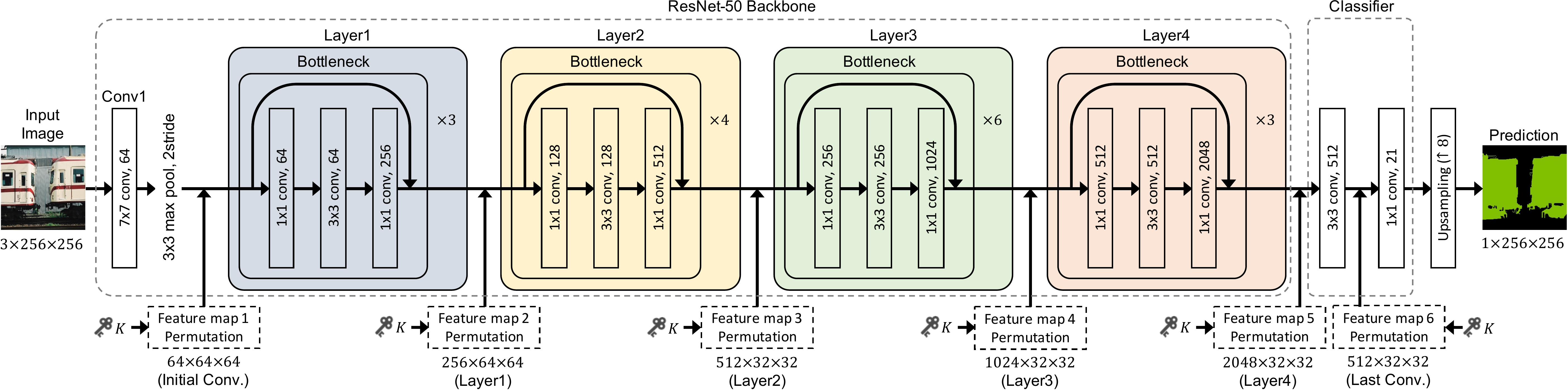}}
  	\caption{Semantic segmentation model (FCN with ResNet-50 backbone) with feature map permutation}
	\label{fig:model}
\end{figure*}

\begin{figure}[t]
	\centering
 	\centerline{\includegraphics[width=\linewidth] {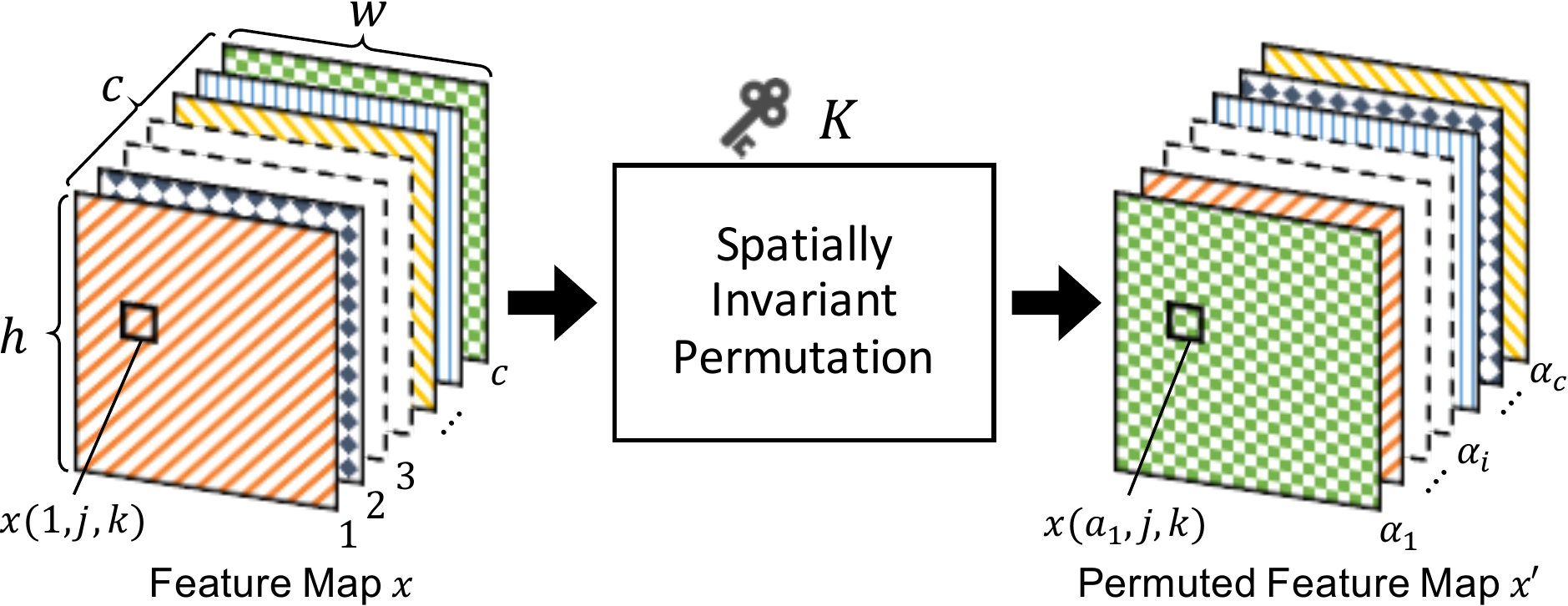}}
  	\caption{Spatially invariant feature map permutation}
	\label{fig:permutation}
\end{figure}

To protect semantic segmentation models, we train models by randomly permuting feature maps with secret key $K$.
The permutation is applied to feature maps in a network as in Fig.~\ref{fig:model}.
In the figure, a fully convolutional network (FCN) \cite{fcn} with a ResNet-50 \cite{resnet} backbone is illustrated as an example, although the permutation is not limited to the FCN.
There are six feature maps in Fig.~\ref{fig:model}, and a number of feature maps from the six maps are chosen to be permuted prior to permutation.
In this paper, a feature map $x$ with a dimension of ($c \times h \times w$), where $c$ is the number of channels, $h$ is the height, and $w$ is the width of the feature map, is transformed with key $K$ at each iteration for training a model.
There are two steps in the process of transforming a feature map as below (see Fig.~\ref{fig:permutation}).
\begin{enumerate}
    \item Generate a random vector with a size of $c$ such that
    \begin{equation}
    [\alpha_1, ., \alpha_i, ., \alpha_{i'}, \ldots, \alpha_c], \alpha_i \in \{ 1, \ldots, c \},
    \end{equation}
    where $\alpha_i \neq \alpha_{i'}$ if $i \neq i'$.
    
    \item Replace all elements of $x$, $x(i, j, k)$, $i \in \{1, \ldots, c\}$, $j \in \{1, \ldots, h\}$, and $k \in \{1, \ldots, w\}$ with $x(a_i, j, k)$ to produce permuted feature map $x'$, where an element of $x'$, $x'(i, j, k)$ is equal to $x(a_i, j, k)$.
\end{enumerate}
If multiple feature maps are chosen to be permuted, the above steps are applied to each feature map.

\begin{figure}[t]
	\centering
 	\centerline{\includegraphics[width=\linewidth] {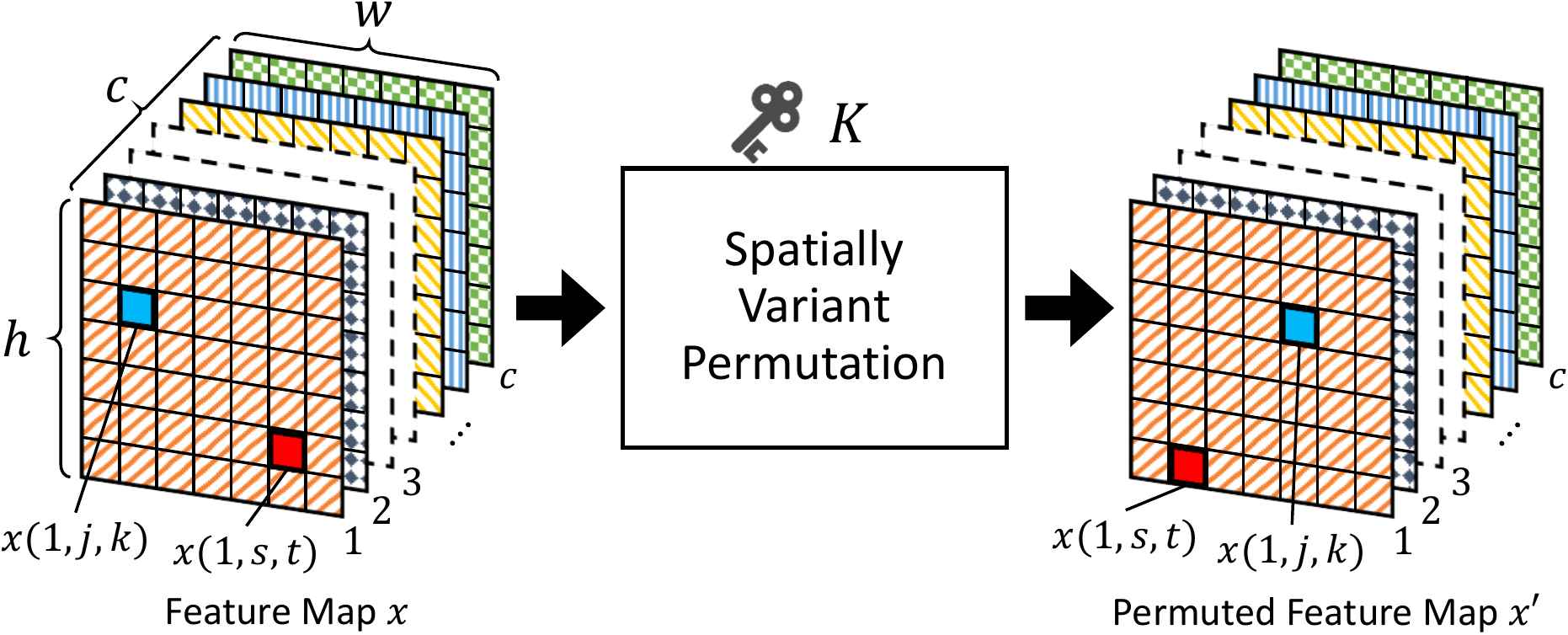}}
  	\caption{Spatially variant feature map permutation}
	\label{fig:transformation}
\end{figure}

The above feature map permutation is spatially invariant as illustrated in Fig.~\ref{fig:permutation}.
This spatially invariant property is important when the permutation is applied to applications that are required to output images like semantic segmentation.
In contrast, an example of spatially variant permutation is given in Fig.~\ref{fig:transformation}.
The conventional protection method for image classification \cite{access3} is carried out by using a spatially variant permutation method, so it is not available for protecting other models such as semantic segmentation models as described later.
In this paper, a model protection method for semantic segmentation is discussed for the first time.

\subsection{Applying Queries to Model}
As shown in Fig.~\ref{fig:framework}, authorized users have key $K$, and key $K$ is also used for semantic segmentation.
In the proposed method, a query image is applied to a model trained with $K$, and the model is protected by permuting feature maps with key $K$ as well as for training the model.
Protected models are expected to satisfy the following requirements.
\begin{enumerate}
    \item Providing almost the same performance as that of using unprotected models to authorized users.
    \item Degrading performance for unauthorized users even when they estimate key $K$.
\end{enumerate}

%% file: sections/experiment.tex
The effectiveness of the proposed method was evaluated in terms of segmentation performance and robustness against unauthorized uses.

\subsection{Experimental Setup}
We used a FCN with a ResNet-50 backbone (see Fig.~\ref{fig:model}) for semantic segmentation.
Segmentation models were trained by using the PASCAL visual object classes dataset released in 2012 \cite{pascal} for semantic segmentation, where the dataset has a training set with 1464 pairs (i.e., images and corresponding ground truths) and a validation set with 1449 pairs.
We split the training set into training and development sets; 1318 samples were used for training, and 146 samples were used for development when training models.
The whole validation set was used for testing the models.
The performance of the trained models was calculated by averaging the mean IoU (see Eq.~(\ref{eq:iou})) values of predictions.
The conventional method using block-wise transformations \cite{access3} requires a fixed input image size.
Therefore, to compare the conventional method with the proposed one, all input images and ground truths were resized to $256 \times 256$.
In addition, standard data-augmentation methods, i.e., random resized crop and horizontal flip, were performed in the training.

All networks were trained for 100 epochs by using the stochastic gradient descent (SGD) optimizer with a weight decay of 0.005 and a momentum of 0.9.
The learning rate was initially set to 0.1 and scheduled by cosine annealing with warm restarts \cite{lr-schedule}.
The minimum learning rate was 0.0001, and the learning rate was restarted every ten epochs.
The batch size was 256.
We used cross-entropy loss to calculate loss.
After the training, we selected the model that provided the lowest loss value under the validation.

\subsection{Performance Evaluation under Correct Key \texorpdfstring{$K$}{K}}
\label{subsec:correct}
In this experiment, one feature map was chosen from six feature maps in the network, and segmentation models were then trained by permuting the chosen feature map with key $K$.
The trained models were evaluated for authorized users with $K$. 
``Correct ($K$)'' in Table~\ref{tab:accuracy} shows the result under this condition, where the model trained by permuting feature map 1 is referred to as Model-1 as an example, and ``Baseline'' denotes that the model was trained and tested without any feature map permutation. 
\begin{table}[t!]
	\centering
	\caption{Segmentation accuracy (mean IoU) of protected models. \\ Best accuracies are shown in bold.}
	\label{tab:accuracy}
	\begin{tabular}{c|ccc} \hline \hline
        Selected feature map & Correct ($K$) & No-perm & Incorrect ($K'$) \\ \hline
        1 (Model-1) & 0.5290 & 0.1678 & 0.0356 \\
        2 (Model-2) & 0.4975 & 0.3735 & 0.0004 \\
        3 (Model-3) & 0.4520 & 0.3799 & 0.0001 \\
        4 (Model-4) & 0.5446 & 0.4040 & 0.0013 \\
        5 (Model-5) & 0.5759 & 0.5328 & \textbf{0.0000} \\
		6 (Model-6) & \textbf{0.5888} & \textbf{0.1595} & \textbf{0.0000} \\ \hline
        Baseline & \multicolumn{3}{c}{0.5966} \\ \hline \hline
	\end{tabular}
\end{table}

From Correct ($K$) in Table~\ref{tab:accuracy}, several models such as Model-6 and Model-5 had a high segmentation accuracy, which was almost the same as that of the baseline, although a couple of models had a slightly degraded accuracy.
Figure~\ref{fig:prediction} also shows an example of prediction results.
From this figure, the proposed model was demonstrated to maintain prediction results similar to the baseline.
\bgroup
\setlength{\tabcolsep}{2pt}
\begin{figure*}[t!]
    \centering
    \begin{tabular}{cc|c|ccc}
    \multirow{2}{*}{Input} & \multirow{2}{*}{Ground Truth} & \multirow{2}{*}{Baseline} &  \multicolumn{3}{c}{Proposed} \\ \cline{4-6}
     & & & Correct & Plain & Incorrect \\ \hline
     & & & & & \\[-8pt]
    \begin{minipage}{2.5cm}
      \centering
      \includegraphics[width=2.5cm]{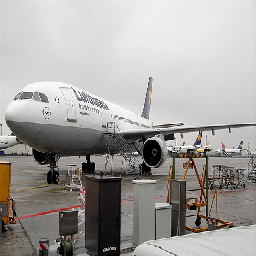}
    \end{minipage} &
    \begin{minipage}{2.5cm}
      \centering
      \includegraphics[width=2.5cm]{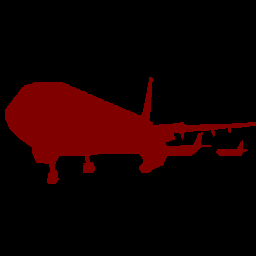}
    \end{minipage} &
    \begin{minipage}{2.5cm}
      \centering
      \includegraphics[width=2.5cm]{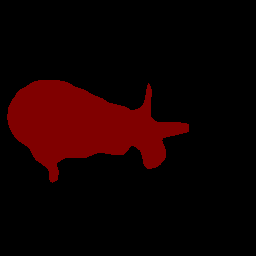}
    \end{minipage} &
    \begin{minipage}{2.5cm}
      \centering
      \includegraphics[width=2.5cm]{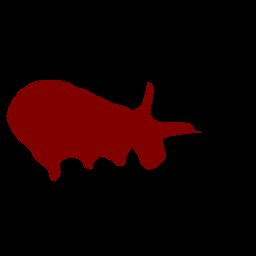}
    \end{minipage} &
    \begin{minipage}{2.5cm}
      \centering
      \includegraphics[width=2.5cm]{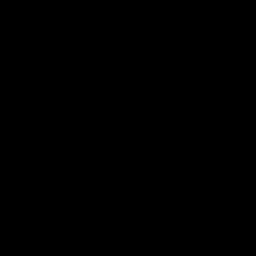}
    \end{minipage} &
    \begin{minipage}{2.5cm}
      \centering
      \includegraphics[width=2.5cm]{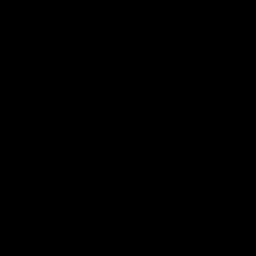}
    \end{minipage} \\ 
     & & 0.8532 & 0.8570 & 0.0000 & 0.0000 \\
    \begin{minipage}{2.5cm}
      \centering
      \includegraphics[width=2.5cm]{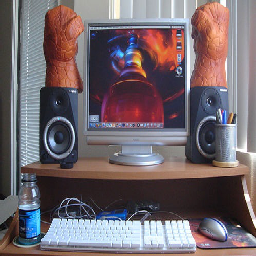}
    \end{minipage} &
    \begin{minipage}{2.5cm}
      \centering
      \includegraphics[width=2.5cm]{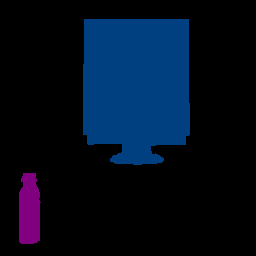}
    \end{minipage} &
    \begin{minipage}{2.5cm}
      \centering
      \includegraphics[width=2.5cm]{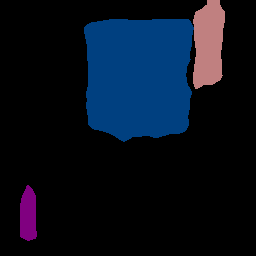}
    \end{minipage} &
    \begin{minipage}{2.5cm}
      \centering
      \includegraphics[width=2.5cm]{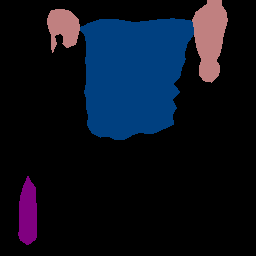}
    \end{minipage} &
    \begin{minipage}{2.5cm}
      \centering
      \includegraphics[width=2.5cm]{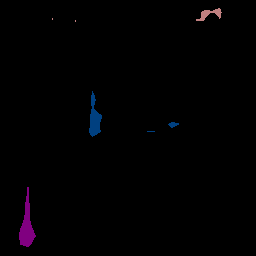}
    \end{minipage} &
    \begin{minipage}{2.5cm}
      \centering
      \includegraphics[width=2.5cm]{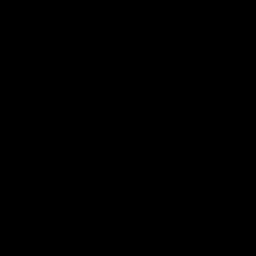}
    \end{minipage} \\
     & & 0.4691 & 0.5062 & 0.1145 & 0.0000 \\
    \begin{minipage}{2.5cm}
      \centering
      \includegraphics[width=2.5cm]{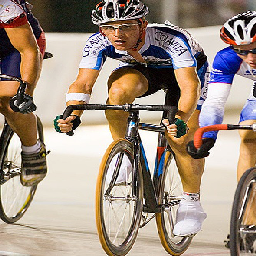}
    \end{minipage} &
    \begin{minipage}{2.5cm}
      \centering
      \includegraphics[width=2.5cm]{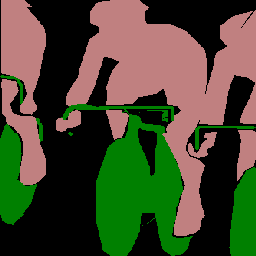}
    \end{minipage} &
    \begin{minipage}{2.5cm}
      \centering
      \includegraphics[width=2.5cm]{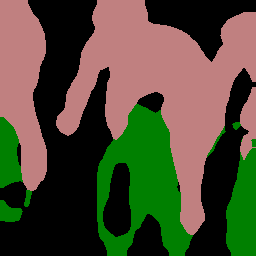}
    \end{minipage} &
    \begin{minipage}{2.5cm}
      \centering
      \includegraphics[width=2.5cm]{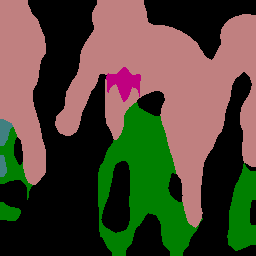}
    \end{minipage} &
    \begin{minipage}{2.5cm}
      \centering
      \includegraphics[width=2.5cm]{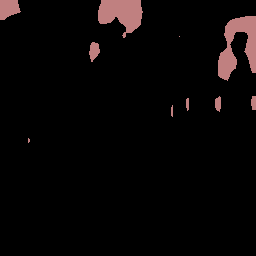}
    \end{minipage} &
    \begin{minipage}{2.5cm}
      \centering
      \includegraphics[width=2.5cm]{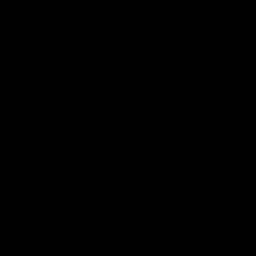}
    \end{minipage} \\
     & & 0.7912 & 0.3679 & 0.0630 & 0.0000 \\
    \begin{minipage}{2.5cm}
      \centering
      \includegraphics[width=2.5cm]{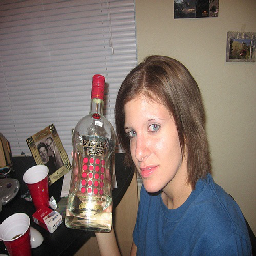}
    \end{minipage} &
    \begin{minipage}{2.5cm}
      \centering
      \includegraphics[width=2.5cm]{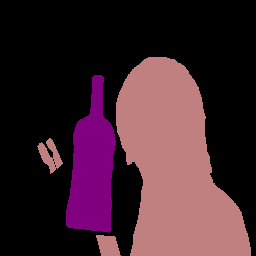}
    \end{minipage} &
    \begin{minipage}{2.5cm}
      \centering
      \includegraphics[width=2.5cm]{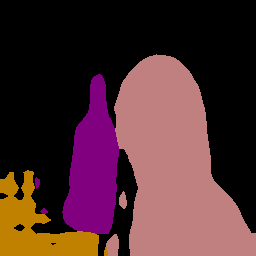}
    \end{minipage} &
    \begin{minipage}{2.5cm}
      \centering
      \includegraphics[width=2.5cm]{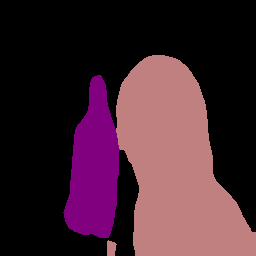}
    \end{minipage} &
    \begin{minipage}{2.5cm}
      \centering
      \includegraphics[width=2.5cm]{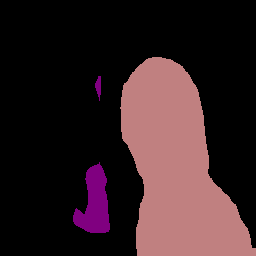}
    \end{minipage} &
    \begin{minipage}{2.5cm}
      \centering
      \includegraphics[width=2.5cm]{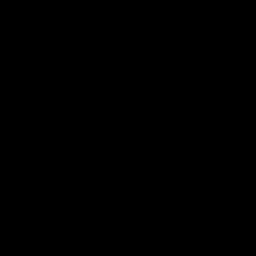}
    \end{minipage} \\
     & & 0.6231 & 0.9279 & 0.6126 & 0.0000 \\
    \end{tabular}
    \caption{Example of prediction results. Mean IoU values are given under predictions.}
    \label{fig:prediction}
\end{figure*}
\egroup

\subsection{Robustness against Unauthorized Access}
We assume that unauthorized users have no key $K$ and that they know both the method for protecting models and the permuted feature maps.
To evaluate robustness against unauthorized access, we also evaluated the six protected models under two key conditions: No-perm and Incorrect, as shown in Table~\ref{tab:accuracy}.
``No-perm'' indicates that protected models were tested without any feature map permutation. 
``Incorrect'' denotes that protected models were tested by permuting a feature map used in training with incorrect (randomly generated) key $K'$.

Table~\ref{tab:accuracy} shows the results under these conditions, where the results for Incorrect were averaged over 100 incorrect keys.
From the table, the protected models provided a low segmentation accuracy, so the proposed models were robust against these attacks.
An example of prediction results is also illustrated in Fig.~\ref{fig:prediction}. 
The robustness of the proposed models can be visually confirmed from the figure as well.

\subsection{Comparison with State-of-the-art Method}
A method for model protection was proposed in \cite{access3}.
In the method, input images are transformed by using three block-wise transformations with a secret key: pixel shuffling (SHF), negative/positive transformation (NP), and format-preserving Feistel-based encryption (FFX) \cite{ffx}.
Although this conventional method can achieve high performance in image classification models, it has never been applied to other models such as semantic segmentation ones.
To be compared with the proposed method, it was also applied to semantic segmentation models.

We trained segmentation models with different block sizes and tested the models with three key conditions, i.e., Correct, Plain, and Incorrect, where ``Plain'' used plain images as input ones, and ``Incorrect'' used images encrypted by using an incorrect key. 
As shown in Table~\ref{tab:conventional}, the performance of all transformations heavily decreased compared with the proposed method.
\begin{table*}[t!]
\centering
\caption{Segmentation accuracy (mean IoU) of transformations of conventional method}
\label{tab:conventional}
\begin{tabular}{c|ccc|ccc|ccc} \hline \hline
    \multirow{2}{*}{Block size} & \multicolumn{3}{c|}{SHF}      & \multicolumn{3}{c|}{NP}       & \multicolumn{3}{c}{FFX}      \\ \cline{2-10}
    & Correct & Plain  & Incorrect & Correct & Plain  & Incorrect & Correct & Plain  & Incorrect \\ \hline
2   & 0.5447 & 0.5112 & 0.4800 & 0.5143 & 0.4370 & 0.0529 & 0.4224 & 0.0047 & 0.0000 \\
4   & 0.4731 & 0.4536 & 0.3671 & 0.4706 & 0.3359 & 0.1505 & 0.3823 & 0.0157 & 0.0012 \\
8   & 0.3599 & 0.3691 & 0.2527 & 0.4094 & 0.3838 & 0.0771 & 0.3613 & 0.0001 & 0.0012 \\
16  & 0.2214 & 0.1994 & 0.1150 & 0.3439 & 0.2114 & 0.0832 & 0.2611 & 0.0007 & 0.0079 \\
32  & 0.1062 & 0.0598 & 0.0398 & 0.2928 & 0.0499 & 0.1239 & 0.1995 & 0.0152 & 0.0770 \\ \hline
Model-6 (Proposed) & \multicolumn{3}{c}{0.5888 (Correct)} & \multicolumn{3}{c}{0.1595 (Plain)} & \multicolumn{3}{c}{0.0000 (Incorrect)} \\ \hline
Baseline (non-protected) & \multicolumn{9}{c}{0.5966} \\ \hline \hline
\end{tabular}
\end{table*}
In particular, when using a large block size, the accuracy for the correct key was low. 
In contrast, the accuracy for the plain and incorrect keys was high, so the performance was confirmed to be poor for protecting semantic segmentation models.

Semantic segmentation models are required to output visual information as an image, so transformations applied to images or feature maps for training and testing models have to be spatially invariant, but the conventional block-wise transformations are not spatially invariant. 
That is why the performance of the block-wise transformations was poor for semantic segmentation models.

%% file: sections/conclusion.tex
We proposed an access control method that uses the spatially invariant permutation of feature maps for protecting semantic segmentation models for the first time. 
Semantic segmentation models are required to output visual information as an image, so transformations for model protection have to be spatially invariant, but conventional transformations are not spatially invariant.
The proposed method allows us not only to obtain a high segmentation accuracy but also for there to be robustness against various attacks by unauthorized users.
In experiments, the effectiveness of the proposed method was verified in terms of segmentation performance and robustness against unauthorized access. 
In contrast, the conventional protection method with block-wise transformations that was proposed for image classification models was demonstrated to not be applicable to segmentation models. 
As future work, we plan to evaluate the robustness against more diverse attacks such as key estimation attacks. 